\documentclass{article}
\usepackage{caption}
\usepackage{spconf,amsmath,graphicx,hyperref}
\usepackage{multirow}
\usepackage{booktabs}
\usepackage{arydshln}
\usepackage{makecell}
\usepackage{array}
\usepackage{adjustbox}
\usepackage{subcaption}
\usepackage{tikz}
\usepackage[table]{xcolor}
\usepackage{colortbl}
\usepackage{pifont}
\usetikzlibrary{arrows.meta}

\title{MTP-S2UT: Enhancing Speech-to-Speech Translation Quality with Multi-token Prediction}

\name{\begin{tabular}{c}
Jianjin Wang$^{1\star}$ \qquad Runsong Zhao$^{1\star}$ \qquad Xiaoqian Liu$^{1}$ \qquad Yuan Ge$^{1}$ \qquad Ziqiang Xu$^{1}$ \qquad \\ Tong Xiao$^{1,2\dagger}$ \qquad Shengxiang Gao$^{3}$ \qquad Zhengtao Yu$^{3}$ \qquad Jingbo Zhu$^{1,2}$\thanks{$\star$ Equal contribution.}\thanks{$\dagger$ Corresponding author.}
\end{tabular}}

\address{$^{1}$ School of Computer Science and Engineering, Northeastern University, Shenyang, China \\
  $^{2}$ NiuTrans Research, Shenyang, China \\
  $^{3}$ Kunming University of Science and Technology, Kunming, China}

\begin{document}
\ninept

\maketitle

\begin{abstract}
Current direct speech-to-speech translation methods predominantly employ speech tokens as intermediate representations. However, a single speech token is not dense in semantics, so we generally need multiple tokens to express a complete semantic unit. To address this limitation, we introduce multi-token prediction (MTP) loss into speech-to-unit translation (S2UT) models, enabling models to predict multiple subsequent tokens at each position, thereby capturing more complete semantics and enhancing information density per position. Initial MTP implementations apply the loss at the final layer, which improves output representation but initiates information enrichment too late. We hypothesize that advancing the information enrichment process to intermediate layers can achieve earlier and more effective enhancement of hidden representation. Consequently, we propose MTP-S2UT loss, applying MTP loss to hidden representation where CTC loss is computed. Experiments demonstrate that all MTP loss variants consistently improve the quality of S2UT translation, with MTP-S2UT achieving the best performance.
\end{abstract}
\section{Introduction}
\label{sec:intro}

Direct speech-to-speech translation~\cite{jia19_interspeech} converts source language speech into target language speech, showing promising applications in international conferences, cross-border communication, and travel scenarios. Recent advances in direct speech-to-speech translation with speech tokens (also known as discrete units) have emerged as a dominant research direction~\cite{lee-etal-2022-direct,inaguma-etal-2023-unity,barrault2023seamlessm4t,zhang-etal-2024-streamspeech}. 

\begin{figure*}[t]
\centering
\captionsetup[sub]{font=small, labelfont=small}
 \begin{minipage}{0.4\textwidth}  
        \centering
        \subcaptionbox{S2UT Model\label{sub1}}{
         \includegraphics{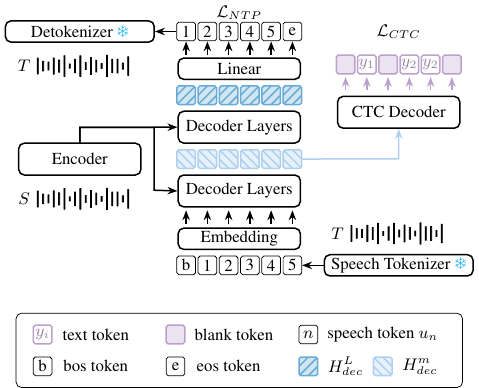}
        }
 \end{minipage}
 \begin{minipage}{0.59\textwidth}  
        \centering
        \begin{minipage}{0.28\textwidth}  
            \centering
            \subcaptionbox{MTP-Parallel-Linear\label{sub2}}{
            \includegraphics{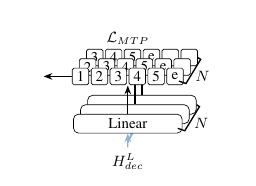}
            }
        \end{minipage}
        \hspace{1cm}
        \begin{minipage}{0.28\textwidth}  
            \centering
            \subcaptionbox{MTP-DeepSeek-V3\label{sub3}}{
            \includegraphics{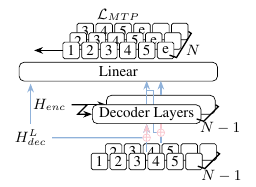}
            }
        \end{minipage}
        \hspace{3cm}
        \begin{minipage}{0.28\textwidth}  
            \centering
            \subcaptionbox{MTP-VocalNet\label{sub4}}{
             \includegraphics{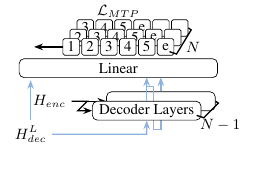}
            }
        \end{minipage}
        \hspace{1cm}
        \begin{minipage}{0.28\textwidth}  
            \centering
            \subcaptionbox{MTP-S2UT\label{sub5}}{
             \includegraphics{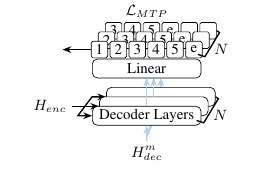}
            }
        \end{minipage}
 \end{minipage}
\caption{Overview of S2UT model and our implementation of 4 MTP loss variants on the S2UT model.}
\label{fig:MTP_Fig}
\end{figure*}

The speech-to-unit translation (S2UT) model~\cite{lee-etal-2022-direct} serves as a representative architecture within this paradigm. As shown in Fig.~\ref{sub1}, it quantizes target speech into discrete speech tokens via a speech tokenizer~\cite{lee-etal-2022-direct,du2024cosyvoice,zeng2024glm}, performs continuous source speech to discrete target speech tokens conversion using sequence-to-sequence models, and synthesizes target speech from these tokens through a detokenizer.

Compared to text tokens, speech tokens exhibit sparse semantic representations, typically requiring multiple speech tokens to express a single semantic concept~\cite{wang2025vocalnet}. This leads to higher prediction entropy and increases modeling complexity~\cite{wang2025vocalnet}. Multi-token prediction (MTP) offers a promising solution to mitigate this challenge by predicting multiple tokens at each position, allowing these tokens to collectively form complete semantic units. MTP~\cite{gloeckle2024better} is originally proposed as an auxiliary task in large language models, as illustrated in Fig.~\ref{sub2}. The approach employs multiple output heads to predict multiple future tokens in parallel, thereby enhancing representational capacity and accelerating inference. Subsequently, DeepSeek-V3~\cite{liu2024deepseek} further refines the MTP module, as shown in Fig.~\ref{sub3}, by introducing additional inputs and Transformer blocks with a focus on improving model performance. VocalNet~\cite{wang2025vocalnet} pioneers the application of MTP in spoken dialogue models~\cite{ji2024wavchat}, as depicted in Fig.~\ref{sub4}. Building upon DeepSeek-V3's MTP design, VocalNet removes the additional inputs to prevent MTP from degenerating into next-token prediction (NTP), thereby effectively alleviating error propagation in speech token prediction while better capturing local patterns~\cite{wang2025vocalnet}.

Our work is the first to introduce the MTP loss variants into the S2UT framework. As shown in Fig.~\ref{Schematic diagram of CTC Shift}, experiments reveal that MTP influences CTC-based text decoding, causing forward shifting of text tokens and backward shifting of blank tokens in CTC alignments, which indicates a forward semantic drift in hidden representations.
Since a semantic unit typically corresponds to multiple speech tokens and each token can only access semantic information from preceding tokens, the early determination of semantic information helps reduce the uncertainty in predicting the corresponding multiple speech tokens.
The hidden layer where CTC loss is computed incorporates both speech and text information and exhibits a significant shifting phenomenon. Therefore, we consider this layer as critical and propose the MTP-S2UT loss, as illustrated in Fig.~\ref{sub5}. We apply the MTP loss to this layer to encourage the model to fuse information from multiple speech tokens and text earlier in the intermediate layers, thereby enhancing the hidden representations in the shallower layers.
Results in Section~\ref{ssec:results} show significant improvements on French$\to$English speech translation across all variants, with MTP-S2UT yielding the most pronounced gains, consistent across speech tokenizers and on Spanish$\to$English tasks. This validates that MTP strengthens the hidden representations of speech tokens, with earlier application leading to superior performance.
Further analyses in Sections~\ref{sssec:ctc shift} and~\ref{sssec:predictive} reveal that MTP loss guides text token forward shifting and reduces the uncertainty in speech token prediction, with this phenomenon being particularly pronounced in MTP-S2UT.

Our work validates the effectiveness of MTP loss within the S2UT framework and demonstrates that applying it to the CTC hidden layer further enhances speech translation performance, providing novel insights into the role of MTP in speech translation.

\section{Method}
\label{sec:method}

This section is structured as follows: Section~\ref{ssec:s2ut} reviews the S2UT model. Section~\ref{ssec:MTPs} details the adaptation of existing MTP losses to S2UT. Our MTP-S2UT loss is finally introduced in Section~\ref{ssec:ourmtp}.

\subsection{S2UT}
\label{ssec:s2ut}


A speech-to-speech translation dataset is typically composed of quadruplets in the form of $D=\{(S, X, Y, T)\}$. Here, $S=(s_1, \cdots, s_{\lvert S \rvert})$ represents the source speech, $X=(x_1, \cdots, x_{\lvert X \rvert})$ is the source text, $Y=(y_1, \cdots, y_{\lvert Y \rvert})$ is the target text, and $T=(t_1, \cdots, t_{\lvert T \rvert})$ is the target speech. As shown in Fig.~\ref{sub1}, in the S2UT~\cite{lee-etal-2022-direct} model, the continuous target speech waveform $T$ is quantized into a sequence of speech tokens $U=(u_1, \cdots, u_{\lvert U \rvert},e)$ by the speech tokenizer, where $e$ denotes the end-of-sequence (eos) token.
The encoder will encode the source speech $S$ into a sequence of hidden states $H_{enc}$.

Subsequently, the encoded source speech representations $H_{enc}$ and right-shifted sequence $U^{+1}=(b, u_1, \cdots, u_{\lvert U \rvert})$ are fed into the decoder to predict $U$ via cross-attention mechanisms, where $b$ denotes the begin-of-sequence (bos) token:
\begin{align}
H_{dec}^0 &= \mathrm{Emb}(U^{+1}) \label{eq:embed} \\
H_{dec}^{i} &= \mathrm{DecoderLayer}^{i}(H_{enc}, H_{dec}^{i-1}) \label{eq:decoder} \\
\mathcal{L}_{NTP} &= -\log P(U|H_{dec}^{L}) \label{eq:NTP} \\
\mathcal{L}_{S2UT} &= \mathcal{L}_{NTP} + \mathcal{L}_{other}
\end{align}
where $L$ is the number of decoder layers and S2UT model employs multi-task training with $\mathcal{L}_{NTP}$ representing the next-token prediction loss, $\mathcal{L}_{other}$ capturing other auxiliary task losses, and $\mathcal{L}_{S2UT}$ as the final combined training objective.

\subsection{Multi-token Prediction}
\label{ssec:MTPs}
Prior works~\cite{wang2025vocalnet,gloeckle2024better,liu2024deepseek} employ MTP loss $\mathcal{L}_{MTP}$ to replace the NTP loss $\mathcal{L}_{NTP}$, applying it exclusively to the final layer hidden representations $H_{dec}^{L}$. Under this paradigm, each token position is required to predict the subsequent $N$ tokens:
\begin{equation}
\mathcal{L}_{MTP} = -\sum_{k=0}^{N-1}\log P(U^{-k}|H_{dec}^{L})
\end{equation}
where $U^{-k}$ denotes the sequence $U$ left-shifted by $k$ positions, with insufficient positions padded accordingly.



The difference between the loss functions of the previous three works lies in the modeling of $P(U^{-k}|H_{dec}^{L})$. As shown in Fig.~\ref{sub2}, \textbf{MTP-Parallel-Linear}~\cite{gloeckle2024better} uses $N$ independent linear heads:

\begin{equation}
P(U^{-k}|H_{dec}^{L}) = \mathrm{softmax}(W^kH_{dec}^{L})
\end{equation}

As shown in Fig.~\ref{sub3}, \textbf{MTP-DeepSeek-V3}~\cite{liu2024deepseek} employs MTP with teacher-forcing and Transformer blocks. Given that S2UT takes additional $H_{enc}$ as input, we implement MTP-DeepSeek-V3 as follows:
\begin{align}
    H_{out}^0 &= H_{dec}^{L} \\
    H_{in}^k &= W_{in}^k[\mathrm{LN}(H_{out}^{k-1});\mathrm{LN}(\mathrm{Emb}(U^{1-k}))]\label{eq:fuse} \\
    H_{out}^k &= \mathrm{Decoder}^k(H_{enc}, H_{in}^k),k>0 \\
    P(U^{-k}|H_{dec}^{L}) &= \mathrm{softmax}(W_{out}H_{out}^k) \label{eq:linear}
\end{align}

In Fig.~\ref{sub3}, the symbol $\oplus$ represents the fusion operation described in Equation~\ref{eq:fuse}, which concatenates the normalized $H_{out}^{k-1}$ and the normalized embedding of $U^{1-k}$ along the feature dimension and reduces the dimension to the original size through a linear layer $W_{in}^k$, where $\mathrm{LN}$ denotes layer normalization, and $[\cdot; \cdot]$ denotes concatenation along the feature dimension. Compared to MTP-Parallel-Linear, MTP-DeepSeek-V3 provides additional $U^{1-k}$ and $H_{enc}$ as inputs, making MTP simpler.

\textbf{VocalNet}~\cite{wang2025vocalnet} argues that using teacher-forcing to input real tokens makes the training objective no different from NTP, failing to alleviate error accumulation and unable to effectively capture local speech structures. Therefore, as shown in Fig.~\ref{sub4}, it removes $\mathrm{Emb}(U^{1-k})$ from MTP-DeepSeek-V3:

\begin{equation}
H_{in}^k = H_{out}^{k-1}
\end{equation}

It is worth noting that although MTP loss introduces many parameters, they can all be discarded during inference, thus not affecting the time and space consumption during inference.

\subsection{MTP-S2UT}
\label{ssec:ourmtp}




To achieve higher translation quality, the S2UT model requires additional text supervision signals to enhance the intermediate representation $H_{dec}^{m}$. Connectionist Temporal Classification (CTC)~\cite{graves2006connectionist} loss $\mathcal{L}_{CTC}$ is typically applied to the decoder intermediate hidden states $H_{dec}^{m}$ to enhance the semantic information of their representations:
\begin{equation}
\mathcal{L}_{CTC} = -\log P(Y|H_{dec}^{m})
\end{equation}
where $1\leq m \leq L$. As shown in Fig.~\ref{sub5}, we believe that $H_{dec}^{m}$ is rich in both textual and speech modal information, therefore applying the MTP loss to $H_{dec}^{m}$ would be more effective:
\begin{equation}
\mathcal{L}_{MTP-S2UT} = -\sum_{k=0}^{N-1}\log P(U^{-k}|H_{dec}^{m})
\end{equation}
The modeling for $P(U^{-k}|H_{dec}^{m})$ is as follows:
\begin{align}
    H_{out}^k &= \mathrm{Decoder}^k(H_{enc}, H_{dec}^{m}) \\
    P(U^{-k}|H_{dec}^{m}) &= \mathrm{softmax}(W_{out}H_{out}^k)
\end{align}
This facilitates earlier integration of cross-modal information, enhancing the semantic density of the intermediate representations.

\section{Experiments}
\label{sec:experiments}

\subsection{Data}
\label{ssec:data}

\subsubsection{Dataset}
\label{sssec:dataset}
We conduct our experiments on the CVSS-C benchmark~\cite{jia-etal-2022-cvss}, a large-scale speech-to-speech translation dataset. We evaluate the MTP loss variants applied to S2UT on the French$\to$English (Fr$\to$En) and Spanish$\to$English (Es$\to$En) translation tasks.

\subsubsection{Pre-processing}
\label{sssec:preprocessing}

For the source speech, 80-dimensional mel-filterbank features~\cite{Povey:192584} are computed, followed by global cepstral mean and variance normalization. For the target speech, three tokenizers are evaluated. The first is an unsupervised tokenizer based on k-means clustering (k=1000) applied to mHuBERT features~\cite{hsu2021hubert}; the resulting discrete units are synthesized using a unit-based vocoder~\cite{lee-etal-2022-direct}. The other two are supervised tokenizers, namely the $\mathcal{S}^3$ tokenizer~\cite{du2024cosyvoice} and the GLM-4-Voice-Tokenizer~\cite{zeng2024glm}, with codebook sizes of 6561 and 16384, respectively. For these tokenizers, speech reconstruction is performed in two stages: a flow matching model~\cite{lipman2023flow} first generates a mel-spectrogram from the tokens, which is then converted to waveform audio by a vocoder~\cite{NEURIPS2020_c5d73680}. Finally, the source and target text are tokenized using SentencePiece~\cite{kudo-richardson-2018-sentencepiece}, resulting in a unigram vocabulary of 6000 tokens for each.

\subsection{Model settings}
\label{ssec:model settings}

The S2UT model employs a 12-layer Conformer~\cite{gulati20_interspeech} encoder with a hidden feature dimension of 256. Two-layer Transformer decoders with identical hidden feature dimensions are connected after the 6th and 8th encoder layers for multi-task learning of source and target language texts, with weights set to 8 for both. The decoder is a Transformer decoder with 6 layers and a hidden feature dimension of 512. A CTC decoder is attached after the 3rd decoder layer for multi-task learning of target language text, with a weight of 1.6.

For the MTP configuration, each speech token predicts the subsequent $N=7$ speech tokens. MTP-Parallel-Linear utilizes $N$ independent linear layers, while the other three variants employ one shared linear layer and multiple independent decoders. MTP-DeepSeek-V3 additionally shares the word embedding layer of the main network. MTP-S2UT applies the MTP loss at the 3rd layer, which is also the same layer where the CTC loss is applied. In preliminary experiments, we observe that increasing the decoder layers of MTP-DeepSeek-V3 from 1 to 3 layers yields a 0.21 improvement in greedy search. For fair comparison, we uniformly set each decoder to 3 layers. The weight for MTP loss is set to 1.0.

\begin{table}[t!]
  \centering
  \footnotesize
  \begin{tabular}{llccc}
  \toprule
    Tokenizer & Model & \makebox[0.5cm][c]{Greedy} & \makebox[0.5cm][c]{Beam5} & \makebox[0.5cm][c]{Beam10} \\

    \specialrule{0em}{1pt}{1pt}
    \midrule
    \multirow{5}{*}{$\mathcal{S}^3$ tokenizer}    
                              & S2UT &  17.79  & 18.98 & 19.15   \\
                              & \;+ MTP-Parallel-Linear &  21.34  & 22.40 & 22.52   \\
                              & \;+ MTP-DeepSeek-V3 &  23.38  & 24.25 & 24.31   \\
                              & \;+ MTP-VocalNet &  23.29  & 24.17 & 24.27   \\
                              & \cellcolor{gray!10}\;+ MTP-S2UT &  \cellcolor{gray!10}\textbf{24.36}  & \cellcolor{gray!10}\cellcolor{gray!10}\textbf{25.14} & \cellcolor{gray!10}\textbf{25.16}   \\

  \specialrule{0em}{1pt}{1pt}
  \cdashline{1-5}
  \specialrule{0em}{1pt}{2pt}

  \multirow{5}{*}{\makecell{HuBERT with\\K-means}} 
                              & S2UT &  22.02  & 23.11 & 23.33   \\
                              & \;+ MTP-Parallel-Linear &  22.03  & 23.07 & 23.10   \\
                              & \;+ MTP-DeepSeek-V3 &  22.73  & 23.86 & 23.87   \\
                              & \;+ MTP-VocalNet &  22.11  & 23.37 & 23.60   \\
                              & \cellcolor{gray!10}\;+ MTP-S2UT &  \cellcolor{gray!10}\textbf{23.59}  & \cellcolor{gray!10}\textbf{24.50} & \cellcolor{gray!10}\textbf{24.53}   \\
  \specialrule{0em}{1pt}{1pt}
  \cdashline{1-5}
  \specialrule{0em}{1pt}{2pt}

  \multirow{5}{*}{\makecell{GLM-4-Voice-\\Tokenizer}} 
                              & S2UT &  21.62  & 23.08 & 23.26   \\
                              & \;+ MTP-Parallel-Linear &  21.92  & 23.36 & 23.56   \\
                              & \;+ MTP-DeepSeek-V3 &  22.99  & 24.27 & 24.45   \\
                              & \;+ MTP-VocalNet &  23.55  & 24.99 & 25.20   \\
                              & \cellcolor{gray!10}\;+ MTP-S2UT &  \cellcolor{gray!10}\textbf{23.97}  & \cellcolor{gray!10}\textbf{25.22} & \cellcolor{gray!10}\textbf{25.26}   \\
  \bottomrule
  \end{tabular}
  \caption{ASR-BLEU scores with three speech tokenizers on CVSS-C Fr$\to$En test set under different MTP loss variants.}
  \label{cvssv_fren}
\end{table}

\subsection{Evaluation}
\label{ssec:evaluation}

We evaluate translation quality using ASR-BLEU. This metric is calculated by first transcribing the synthesized target speech into text with an Automatic Speech Recognition (ASR) model, and then computing the BLEU score between the transcription and the ground-truth reference text. For this purpose, we use the oct22 version of the ASR model in the fairseq~\cite{ott2019fairseq} toolkit.

\begin{table}[t!]
  \centering
  \footnotesize
  \begin{tabular}{lccc}
  \toprule
    Model & \makebox[1.3cm][c]{Greedy} & \makebox[1.3cm][c]{Beam5} & \makebox[1.3cm][c]{Beam10} \\

    \specialrule{0em}{1pt}{1pt}
    \midrule
    \multirow{1}{*}{S2UT}   &  16.67  & 17.99 & 18.18   \\


   \multirow{1}{*}{\makecell{\;+ MTP-Parallel-Linear}} 
                              &  16.83  & 18.35 & 18.58   \\
   \multirow{1}{*}{\makecell{\;+ MTP-DeepSeek-V3}} 
                              &  18.94  & 20.14 & 20.31   \\
   \multirow{1}{*}{\makecell{\;+ MTP-VocalNet}} 
                              &  19.98  & 21.47 & 21.69   \\
   \multirow{1}{*}{\cellcolor{gray!10}\makecell{\;+ MTP-S2UT}} 
                              &  \cellcolor{gray!10}\textbf{21.87}  & \cellcolor{gray!10}\textbf{22.59} &\cellcolor{gray!10}\textbf{22.83}   \\
                              
  \bottomrule
  \end{tabular}
  \caption{ASR-BLEU scores with $\mathcal{S}^3$ tokenizer on CVSS-C Es$\to$En test set under different MTP loss variants.}
  \label{cvssv_esen}
\end{table}

\subsection{Main Results}
\label{ssec:results}

We train S2UT models on the CVSS-C Fr$\to$En dataset using four MTP loss variants. Table~\ref{cvssv_fren} shows that models trained with our MTP-S2UT loss variant achieve the best results across all tokenizers and decoding methods. With the $\mathcal{S}^3$ tokenizer and greedy search, the ASR-BLEU score of S2UT improves from 17.79 to 24.36. Other MTP loss variants also bring gains, though smaller ones. This proves that MTP loss boosts the information in hidden states, which leads to better translation. The effect works best when we boost earlier hidden states.

We also test on the CVSS-C Es$\to$En dataset. Table~\ref{cvssv_esen} shows results that match our Fr$\to$En experiments, proving that MTP loss brings gains across different languages.

\subsection{Analysis}
\label{ssec:analysis}

\begin{figure}[t]
\centering
\centerline{\includegraphics{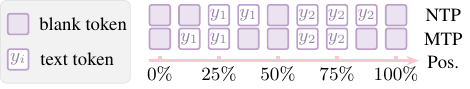}}
\caption{Example of CTC output sequences decoded from the intermediate hidden states $H^m_{dec}$. Compared to NTP loss, the model trained with MTP loss produces text tokens $y_1$ that exhibit an overall forward shift. After each text token's first occurrence, subsequent tokens can access its semantic information, thus we use the first occurrence position to represent the earliest availability of its semantic information. For instance, the first occurrence positions of $y_1$ and $y_2$ from the model  trained with MTP loss are 12.5\% and 62.5\%, respectively. Complete statistics are provided in Table~\ref{cvssv_ctc}.}
\label{Schematic diagram of CTC Shift}
\end{figure}

\begin{table}[t!]
  \centering
  \footnotesize
  \begin{tabular}{lccc}
  \toprule
        Model & \makebox[1.4cm][c]{$\mathcal{S}^3$ tokenizer} & \makebox[1.4cm][c]{\makecell{HuBERT with\\K-means}} & \makebox[1.4cm][c]{\makecell{GLM-4-Voice-\\Tokenizer}} \\

    \specialrule{0em}{1pt}{1pt}
    \midrule
    \multirow{1}{*}{S2UT}   &  51.011\%  & 49.628\% & 50.363\%   \\


   \multirow{1}{*}{\makecell{\;+ MTP-Parallel-Linear}} 
                              &  47.725\%  & 47.288\% & 49.601\%   \\
   \multirow{1}{*}{\makecell{\;+ MTP-DeepSeek-V3}} 
                              &  50.166\%  & 49.267\% & 50.719\%   \\
   \multirow{1}{*}{\makecell{\;+ MTP-VocalNet}} 
                              &  47.504\%  & 42.486\% & 48.889\%   \\
   \multirow{1}{*}{\cellcolor{gray!10}\makecell{\;+ MTP-S2UT}} 
                              &  \cellcolor{gray!10}47.382\%  & \cellcolor{gray!10}44.561\% & \cellcolor{gray!10}43.889\%   \\
                              
  \bottomrule
  \end{tabular}
  \caption{Average relative position of the first occurrence of text tokens. If the average position is $>50\%$, it indicates that text tokens generally appear later in the sequence relative to blank tokens.}
  \label{cvssv_ctc}
\end{table}

\subsubsection{CTC Decoding Forward Shift}
\label{sssec:ctc shift}

In theory, MTP loss encourages early planning of future information. We can verify this hypothesis by performing CTC decoding on the intermediate hidden states $H^m_{dec}$. As shown in Fig.~\ref{Schematic diagram of CTC Shift}, models trained with MTP loss exhibit a forward shift in text tokens compared to those trained with NTP loss, indicating that the semantic information in hidden states is advanced along the sequence dimension. 

To quantify this advancement, we calculate the relative position of the first occurrence of all text tokens within the entire sequence.  We compute the average position of all text tokens, with results presented in Table~\ref{cvssv_ctc}. The results demonstrate that MTP loss significantly shifts text tokens forward, except for MTP-DeepSeek-V3. This exception occurs because MTP-DeepSeek-V3 utilizes additional input through teacher forcing, providing an extra information source that maintains relatively stable semantics. In contrast, other MTP loss variants drive semantic information to shift forward appropriately to achieve future token prediction.

We hypothesize that the forward shift of semantic information can reduce the model's uncertainty in predicting speech tokens to some extent. 
For example,
\textit{Hello} occupies one text token and multiple speech tokens. If the semantic representation of \textit{Hello} appears at earlier positions among these speech tokens, the model can easily predict the speech tokens corresponding to \textit{Hello}. Conversely, if the semantic representation of \textit{Hello} appears at later positions among these speech tokens, the model exhibits higher uncertainty for the initial speech tokens, as it lacks knowledge about which semantic speech tokens to predict. We will measure the model's uncertainty regarding speech tokens in the next subsection.

\subsubsection{Speech Token Uncertainty}
\label{sssec:predictive}

We employ entropy to measure the uncertainty of the model's predictions for speech tokens:

\begin{equation}
\mathrm{Entropy}(\mathrm{Token}) = -\sum_{z \in \mathcal{Z}} p(z) \log p(z)
\end{equation}
where $\mathcal{Z}$ is the set of all possible tokens, and $p(z)$ is the probability that the token is $z$.

\begin{figure}[t]
\centering
\centerline{\includegraphics[width=8.5cm]{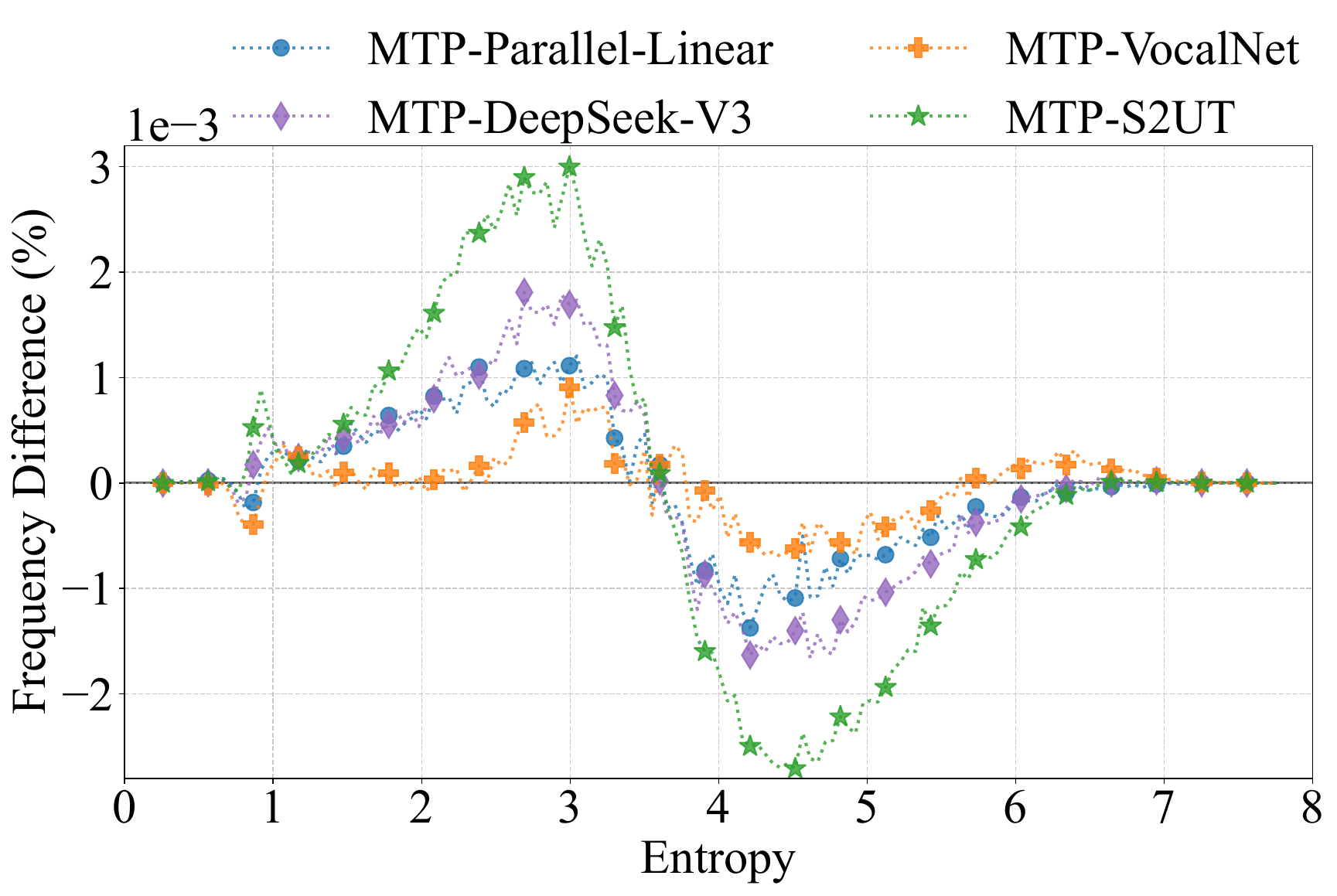}}
\caption{Entropy distribution of 1.2M speech token predictions in S2UT models trained with MTP loss. Speech tokens are from the CVSS-C Fr$\to$En test set using the $\mathcal{S}^3$ tokenizer. All frequencies are presented relative to the baseline (with NTP loss) for enhanced visualization clarity.}
\label{fig:entropy}
\end{figure}

We compute the entropy of five models over 1.2 million tokens using the $\mathcal{S}^3$ tokenizer on the CVSS-C Fr$\to$En test set. To better compare the impact of MTP loss on S2UT models, we subtract the frequency distribution of the S2UT baseline from the frequency distributions of the four MTP loss variants. As shown in Fig.~\ref{fig:entropy}, all MTP loss variants lead to increased frequency in low-entropy regions and decreased frequency in high-entropy regions for S2UT models, indicating that MTP loss indeed reduces uncertainty during the prediction process. Among these variants, our MTP-S2UT demonstrates the most significant and pronounced reduction in uncertainty.

\section{Conclusions}
\label{sec:Conclusions}

This paper introduces multi-token prediction (MTP) into the S2UT framework and proposes a novel MTP-S2UT loss applied at the intermediate CTC layer, significantly enhancing translation quality by encouraging earlier and richer fusion of semantic information across speech and text modalities. Experimental results on Fr$\to$En and Es$\to$En tasks demonstrate consistent and substantial improvements across multiple speech tokenizers and decoding strategies, with MTP-S2UT achieving the highest gains. Our analysis reveals that MTP not only reduces predictive uncertainty for speech tokens but also induces a forward shift in CTC alignments, indicating more efficient semantic planning. This work validates the effectiveness of MTP in speech-to-speech translation and highlights the importance of early intermediate layer enrichment, paving the way for more powerful and efficient direct S2UT models in the future.




\clearpage
\ninept
\bibliographystyle{IEEEbib}
\bibliography{strings,refs}

\end{document}